\documentclass[10pt,twocolumn,letterpaper]{article}

\usepackage{wacv}              

\usepackage{graphicx}
\usepackage{amsmath}
\usepackage{pifont}
\usepackage{amssymb}
\usepackage{booktabs}
\usepackage{multirow}
\usepackage{soul}
\usepackage{enumitem}
\usepackage[pagebackref,breaklinks,colorlinks]{hyperref}

\usepackage[capitalize]{cleveref}
\crefname{section}{Sec.}{Secs.}
\Crefname{section}{Section}{Sections}
\Crefname{table}{Table}{Tables}
\crefname{table}{Tab.}{Tabs.}

\def\wacvPaperID{2694} 
\def\confName{WACV}
\def\confYear{2024}

\begin{document}
{
\title{ZEETAD: Adapting Pretrained Vision-Language Model for \\ 
Zero-Shot End-to-End Temporal Action Detection}

\author{Thinh Phan$^{\ast}$, Khoa Vo$^{\ast}$, Duy Le$^{\dagger}$, Gianfranco Doretto$^{\ddagger}$, Donald Adjeroh$^{\ddagger}$, and Ngan Le$^{\ast}$ \\
$^{\ast}$AICV Lab, University of Arkansas, Fayetteville, Arkansas, USA \\
$^{\dagger}$FPT Software AI Center, Vietnam \\
$^{\ddagger}$West Virginia University, Morgantown, West Virginia, USA \\
{\tt\small \{thinhp, khoavoho, thile\}@uark.edu}, {\tt\small duylda1@fpt.com} \\ 
{\tt\small\{gianfranco.doretto, donald.adjeroh\}@mail.wvu.edu}
}

}
\maketitle

\begin{abstract}

Temporal action detection (TAD) involves the localization and classification of action instances within untrimmed videos. While standard TAD follows fully supervised learning with closed-set setting on large training data, recent zero-shot TAD methods showcase the promising open-set setting by leveraging large-scale contrastive visual-language (ViL) pretrained models. However, existing zero-shot TAD methods have limitations on how to properly construct the strong relationship between two interdependent tasks of localization and classification and adapt ViL model to video understanding. In this work, we present ZEETAD, featuring two modules: dual-localization and zero-shot proposal classification. The former is a Transformer-based module that detects action events while selectively collecting crucial semantic embeddings for later recognition. The latter one, CLIP-based module, generates semantic embeddings from text and frame inputs for each temporal unit. Additionally, we enhance discriminative capability on unseen classes by minimally updating the frozen CLIP encoder with lightweight adapters. Extensive experiments on THUMOS14 and ActivityNet-1.3 datasets demonstrate our approach's superior performance in zero-shot TAD and effective knowledge transfer from ViL models to unseen action categories. Code is available at \url{https://github.com/UARK-AICV/ZEETAD}.

\end{abstract}

\section{Introduction}

With the rapid growth of video content on the internet and social media, video understanding, which is about analyzing and interpreting action sequences, has gained a lot of interest. While video action recognition requires categorizing a standardized snippet with a single label, temporal action detection (TAD) aims to both localize and classify every action instances from long untrimmed videos. This task is challenging because existing supervised methods needs training with large amount of video data to attain decent performance. At the same time, obtaining multiple annotation pairs of temporal regions and corresponding action labels per video is laborious and expensive. These issues restrict current TAD works to closed-set learning setting, where the same set of categories apply to training and inference stage. Hence, there has been an increasing demand for expanding TAD methods to unseen classes with little additional annotation cost via few-shot or preferably zero-shot (ZS) learning strategies.

\begin{figure*}[t]
\centering
\includegraphics[width=1.8\columnwidth]{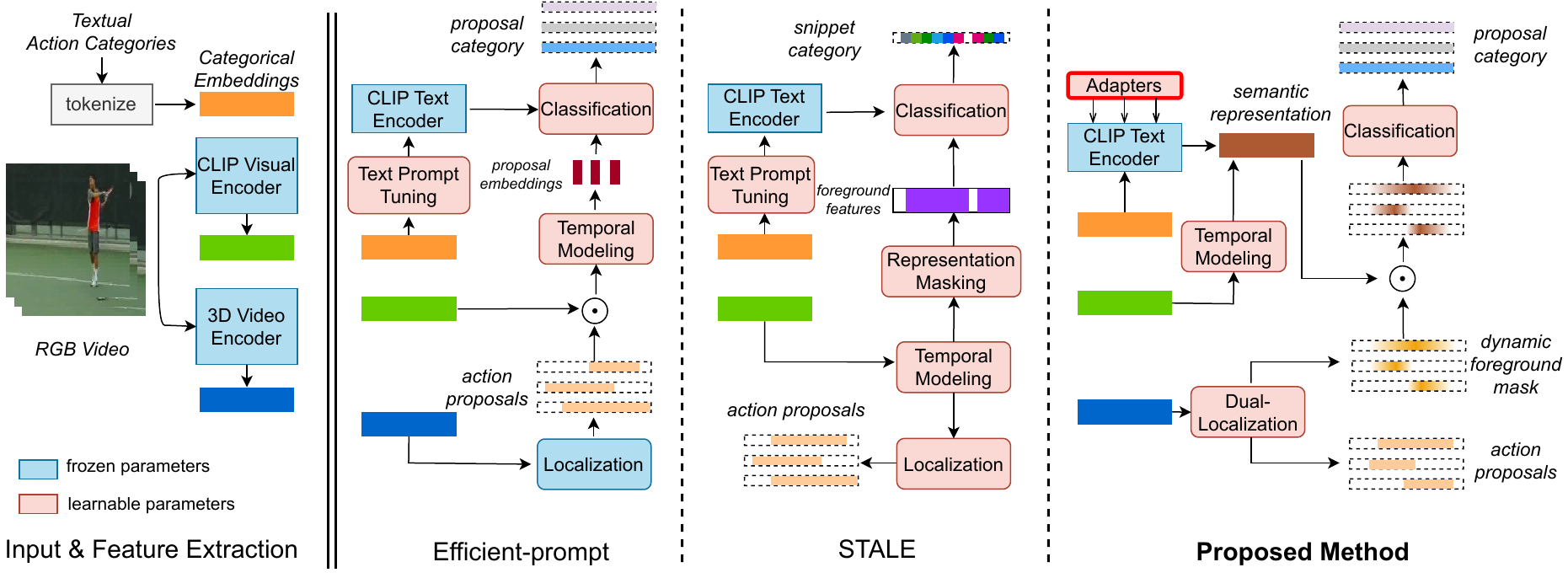}
\caption{Network comparison between our ZEETAD and Efficient-Prompt \cite{ju2022prompting}, STALE \cite{nag2022zero}. Efficient-Prompt acquired the class-agnostic action proposals from pretrained action localization model and applied ZS video action recognition model on them. STALE used CLIP visual encoder for localization instead of 3D video backbone and predicted the action category and event duration on every snippet features. In our approach, we fuse CLIP text and visual features into temporal semantic representation, which we then correlate with a dynamic foreground mask to facilitate subsequent classification. Moreover, to enhance transferability, our technique employs Adapters in lieu of Text Prompt Tuning as in existing works.}
\label{fig:compare}
\end{figure*}

The recent achievements on vision-language (ViL) pretrained models with representatives such as CLIP \cite{radford2021learning}, ALIGN \cite{jia2021scaling}, UniCL\cite{yang2022unified}, Grounding DINO \cite{liu2023grounding}, have not only been beneficial to ZS image understanding tasks \cite{gao2021clip, zhang2021tip, cai2022x, zou2023generalized, liu2022open, ding2022decoupling, nguyen2023language, pham2023decoding} but promoted the generalizability of video analysis issues \cite{lin2022frozen, wu2023bidirectional, joo2023clip, tran2023z, yamazaki2022vlcap, yamazaki2023vltint}. Their success is ascribed to the rich semantics aligned with strong visual representation acquired from web-scale image-text pairs. The ZS transferability is accomplished by matching the similarity between query text embeddings and novel image features, or vice versa. Based on pretrained ViL models, there have been many literature that centered around ZS image recognition \cite{gao2021clip, zhang2021tip, nguyen2023language}, open-vocabulary image segmentation \cite{cai2022x, zou2023generalized, liu2022open, ding2022decoupling}, interpretable AI in medical \cite{nguyen2023language} and some related to ZS action recognition \cite{lin2022frozen, wu2023bidirectional, joo2023clip}, video captioning \cite{yamazaki2022vlcap, yamazaki2023vltint}, objects tracking \cite{tran2023z}. ZS TAD, which is the extension of ZS video action detection, has recently also received more attention and demonstrated promising performance in detecting actions within open-set settings.


As a foundational study, Efficient-Prompt \cite{ju2022prompting} focuses on enhancing CLIP's text encoder to maximize similarity between visual proposals and textual embeddings. 
To achieve this, frame embeddings extracted from CLIP visual encoder are passed through a lightweight Transformer \cite{vaswani2017attention}, capturing temporal information and encapsulating them into proposal embeddings. Nonetheless, due to the utilization of independently obtained proposal boundaries from a pretrained action localization model, the framework lacks integration between localization and classification \cite{nag2022zero} (limitation 1). 
Recognizing this limitation, STALE \cite{nag2022zero} designed a one-stage TAD model that reserved the foreground features through representation masking and jointly accommodated and classified every snippet embeddings. Additionally, the authors adopted text prompt tuning (TPT), a finetuning technique on CLIP, to determine optimal textual action descriptions. They also introduced a cross-modal adaptation to guide text features using contextual-level information.
However, there is a dilemma of the localization input (limitation 2). According to CLIP's formation, novel classes are distinguished by comparing the similarity among CLIP-encoded visual and text embeddings. In STALE, the action localizer receives CLIP-encoded frame features as input. This impedes localization, as these features lack temporal relations and motion cues. Despite the subsequent temporal modeling module, the improvement remains marginal. Replacing this with conventional 3D video encoders for better localization results in an impaired action classifier, given the incongruity between visual features and text embeddings. 
Another issue is that STALE classified every video frames, instead of sequence of frames, and tied them to action proposals. Regarding the activity with several sub-actions, local temporal feature cannot be representative of an entire action (limitation 3). 

Both STALE and Efficient-Prompt adopt ViL model leveraging learnable textual tokens from CLIP's text encoder to enhance ViL model adaptation. However, as noted in \cite{zang2022unified}, TPT struggles to generalize to new classes and grapples with high intra-class variance in visual features (limitation 4).
\textit{To address the aforementioned issues, in this work, we design an effective end-to-end model architecture with the aid of CLIP pretrained model for ZS TAD called ZEETAD}. The network differences between the existing methods and ours are highlighted in Fig.\ref{fig:compare}.

Our method focuses on solving two primary concerns: i) enhancing the ViL model for novel action detection; (ii) integrating the action localizer and classifier within the framework of TAD for an open-set scenario.
Our network is a one-stage TAD model with learnable dual-localization module and ZS proposal classification module to address limitation 1. To rectify the limitation 2 identified in STALE, snippet features extracted from pretrained 3D convolutional neural network (CNN) are used for localization module while frame embeddings from CLIP's image encoder are allocated to the classification module, aligning with their respective objectives. Inspired by semantic image segmentation, the localization module employs video semantic embeddings generated by the classifier module to ascertain regions relevant to the action. Consequently, based on this segmented data, action proposals from seen or unseen classes are determined. 
Specifically, the frame embeddings, after being modeled with temporal relationship, are combined with their counterpart ones to generate a semantic representation that describe the action probabilities for each frames. 
Recognizing that not all frames contribute equally to action discrimination, we introduce the \textit{dual-localization} module, facilitating the gathering of exclusively action-relevant embeddings. The classification of a class is attained by combining selected embeddings and identifying the category with the highest matching score.
Moreover, to achieve better transferability on video domain, we opt for an efficient finetuning technique known as AdaptFormer. As shown in \cite{chen2022adaptformer}, AdaptFormer merely updates the lightweight adapters injected inside the frozen CLIP Transformer sub-layers, however, it surpasses the full-tuning solution by approximately 10\% in video recognition task. The selection of the aforementioned components offers a simple pipeline as well as low computation cost but yields remarkable performance.
 
The main contributions are summarized as follows:
\begin{itemize}[noitemsep]
\item We introduce a dual-localization mechanism designed not only to determine proposal boundaries but also to segment the semantic embeddings synthesized from CLIP. 
\item We integrate an efficient finetuning method, known as Adapters, to adapt a large-scale ViL model to the video domain.
\item As a result, we employ the dual-localization mechanism and Adapters to propose a highly effective end-to-end model architecture for Zero-shot Temporal Action Detection (ZEETAD), encompassing two modules: temporal action dual-localization and Zero-Shot (ZS) proposal classification.
\item We conduct experiments on the THUMOS14 and ActivityNet-1.3 datasets. Our ZEETAD model, featuring dual-localization and conceptual-based classification, significantly outperforms other state-of-the-art (SOTA) methods. We also present comprehensive ablation studies to demonstrate the effectiveness of each individual component. 

\end{itemize}

\section{Related Works}

\subsection{Pretrained Vision-Language Models} 

Vision-Language models have rapidly evolved in recent years with the intention of improving vision models' generalizability upon unseen object classes. The key idea is to capitalize on large scale of pairs of images and natural language descriptions and then train a network to align the image representation with text embeddings through noise contrastive learning. While early approaches explored the semantic representation through word embeddings of class name \cite{al2016recovering} or attributes \cite{lampert2013attribute}, recent works with representatives as CLIP \cite{radford2021learning} and ALIGN \cite{jia2021scaling} augmented the training procedure with millions of image-text pairs as well as the backbone with modern Transformer \cite{vaswani2017attention}. Their rich vision-language correspondence knowledge serves as effective pretrained model for many few-shot to ZS tasks such as image captioning \cite{mokady2021clipcap, yamazaki2022vlcap, yamazaki2023vltint}, image retrieval \cite{liu2021image}, semantic segmentation \cite{ding2022decoupling, tran2022aisformer}, medical imaging \cite{nguyen2023embryosformer, pham2023decoding}, object tracking \cite{li2023ovtrack, tran2023z} . Along with that, adaptation methods \cite{lester2021power, jia2022visual, jie2022convolutional} for large-scale ViL models have been becoming favored research which is about minimally finetuning these computation-heavy models but still boost the generalization capability on new tasks. In this work, we utilize CLIP for ZS action classification branch in TAD and further improve its performance on unseen video categories by incorporating adapters \cite{chen2022adaptformer} to the text encoder backbone.

\begin{figure*}[t]
\centering
\includegraphics[width=2.0\columnwidth]{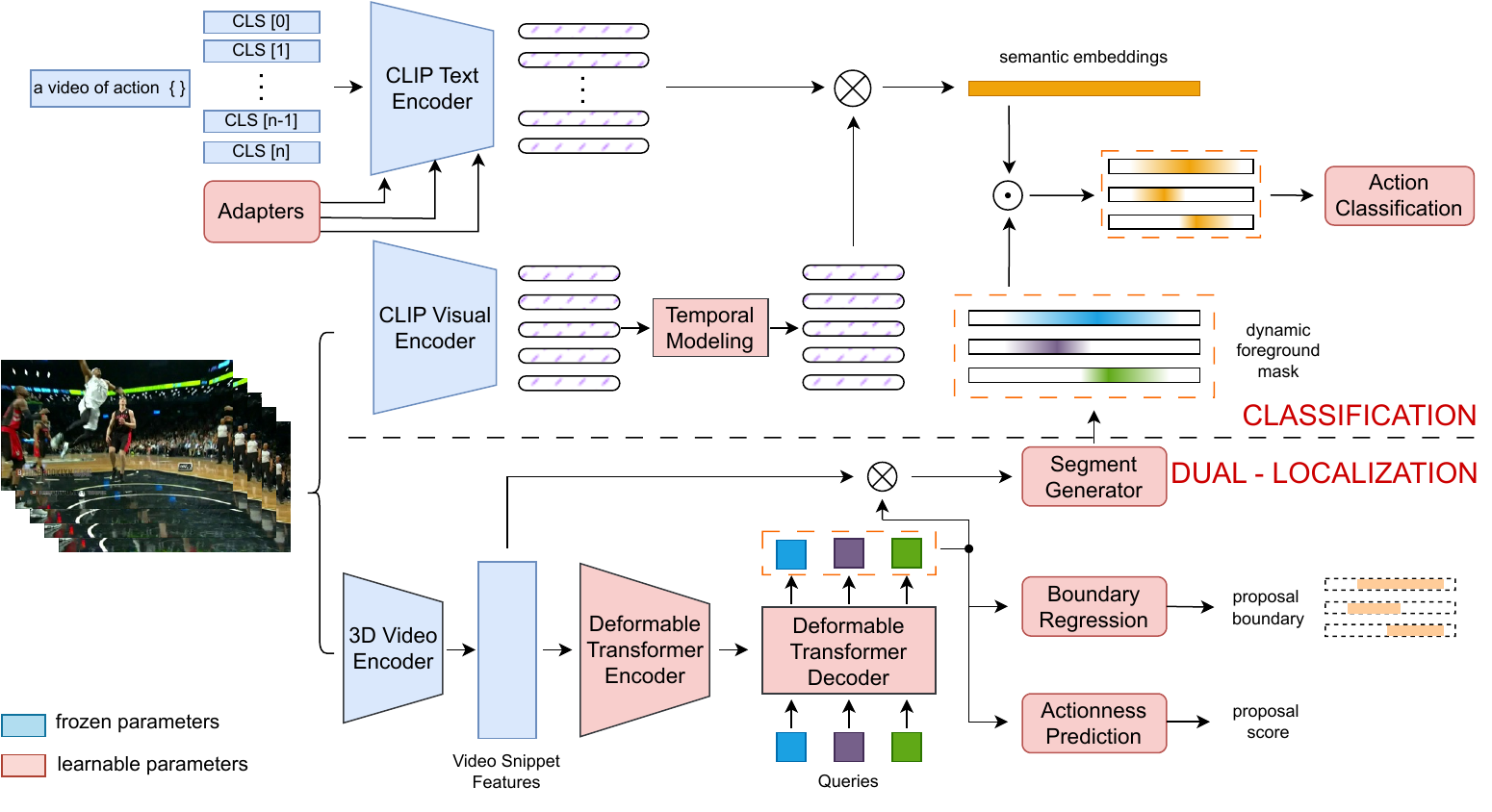}
\caption{Overall architecture of ZEETAD. The classification module associates frame embeddings and textual categories embeddings from CLIP encoders to generate the temporal semantic embeddings. The dual-localization receives the 3D encoded video snippet features and passes them through Deformable Transformer for class-agnostic action proposal generation. Segment generator is proposed in the dual-localization module to dynamically mask the appropriate semantic embeddings for each event. Finally, the selected proposal segment of semantic representation are aggregated and classified. Adapters as a finetuning technique are attached to CLIP text encoder to increase the generalizability of zero-shot classification.}
\label{fig:architecture}
\end{figure*}

\subsection{Temporal Action Detection}
Temporal action detection is one of the key task in video understanding topic. Current methods could be roughly categorized into two-stage methods and one-stage methods. The former ones initially generate action proposals or foreground instances and then assign them with action categories. Commonly, more effort was put on the first stage and action labels were obtained from external classification scores. DCAN \cite{chen2022dcan} followed the anchor-based localization, adjusting the pre-defined anchors based on boundary level and proposal level scores. BSN \cite{lin2018bsn} and BMN \cite{lin2019bmn} fall into action-guided localization direction, evaluating candidate proposals with probability of being a potential action. ABN \cite{vo2021abn} and \cite{vo2021agent} formulates TAD as a interaction between environment and agent. AEI \cite{vo2021aei}, \cite{vo2022contextual}, AOE-Net\cite{vo2023aoe} are also two stages TAD but they are are designed with explainable capability. Single-stage method \cite{zhang2022actionformer, lin2021learning}, also known as anchor-free localization method, simultaneously classifies every temporal units and regresses their boundaries. Our proposed model follows the single-stage method but it concurrently generates the action proposals and categorizes them. A Transformer encoder-decoder as the backbone is responsible for action boundary detection and proposal ranking through actionness score.

\subsection{Zero-shot Temporal Action Detection}
Zero-shot learning considers the model awareness of novel classes absent during training. Zero-shot learning in TAD is challenging because it deals with the joint localization and classification of multiple unseen instances. ZSTAD \cite{zhang2020zstad} adopted the R-C3D framework and optimized activity label mapping by considering common semantics between seen and unseen activities. However, ZSTAD allowed the unseen semantic embeddings to support the training stage via super-class classification loss, which is impractical in real-world scenarios. To address this issue, TranZAD \cite{nag2023semantics} used semantic information of only seen classes at training phase and also proposed a network that learns to group action visual features and their corresponding class-specific semantic embedding. The semantic embeddings in these two methods are acquired from Word2Vec \cite{goldberg2014word2vec} or GLoVe \cite{pennington2014glove}. More recent approaches, such as Efficient-Prompt \cite{ju2022prompting} and STALE \cite{nag2022zero}, harness large-scale pretrained ViL models, which inherently possess visual-textual alignment capabilities. he experiments clearly illustrate the effectiveness of incorporating ViL capabilities to address agnostic-action scenarios in ZS TAD. Both Efficient-Prompt and STALE have outperformed all existing ZS TAD methods. In our approach, we harness the power of the CLIP pretrained model within the action classifier module to accurately identify unseen classes. Specifically, we adapt CLIP to produce temporal semantic embeddings, which are subsequently clustered for each action candidate and ultimately result in the assignment of the final action class.

\section{Methodology}
As shown in Fig.\ref{fig:architecture}, our framework is the unification of two sub-tasks, action proposal localization and action classification. In the upper part of Fig.\ref{fig:architecture}, CLIP pretrained model encodes the RGB frames and action category contents and the frame and text embeddings are subsequently multiplied to create semantic embeddings for all frames. A deformable Transformer encoder-decoder model receives the 3D encoded video features, predicting the action intervals and confidence scores. Unlike supervised TAD that assigns labels by distinguishing video features, ZS TAD indirectly classifies action by assessing the matching score of visual and textual embeddings. Hence, segment generator is designed to generate foreground mask of the semantic embeddings pertinent to the corresponding action boundary. Finally, the assembled semantic embeddings are fed to an classifier to yield the activity label.

\subsection{Problem Definition} \label{ssec:num1}
Our work focuses on the problem of ZS TAD. Given a training set of untrimmed videos $\mathcal{D}_{train} = \{\mathcal{V}_{i}\}_{i=1}^{n}$, we have an input set of RGB frames $X_{i} = \{f_t\}_{t=1}^{T}$, where $\mathcal{T}$ is the number of video snippets from input sequence. As a TAD task, annotation is demonstrated as $Y_i = \{s_k, e_k, {c_s}_{k}\}_{k=1}^{K_i}$ 
where ${K}_i$ is the amount of action events, $s$ and $e$ are respectively the start and end of each event and $c_s$ is activity category. The testing set $\mathcal{D}_{test}$ shares the same data structure of $\mathcal{D}_{train}$; regarding the ZS scenarios, the activity classes in $\mathcal{D}_{train}$ and $\mathcal{D}_{test}$ are non-overlapping $\mathcal{C}_{train} \cap \mathcal{C}_{test} = \emptyset $. 
Our proposed method aims to predict all video segments in an open-set scenario guided by word embeddings of $\mathcal{C}_{test}$.

\subsection{Semantic Representation} \label{ssec:num2}
We adopt CLIP \cite{radford2021learning} pretrained Vision-Language model to set up the ZS action recognition module. To fit CLIP into ZS TAD, our idea is to fuse each frame embedding with entire set of textual category embeddings, creating the semantic embedding for every frames. The semantic representation brings the probabilities of every classes at each temporal unit. Multiple action proposals in a video input are assigned with labels via segmenting and aggregating related semantic embeddings. Specifically, video frame embeddings $\mathcal{F}^{rgb} \in \mathbb{R}^{T \times d}$ are acquired by passing middle RGB frames of video snippets through CLIP visual encoder $\Phi_{CLIP-v}(.)$. To inject the temporal context into unattached frame embeddings, we apply a temporal Transformer \cite{vaswani2017attention} on $\mathcal{F}^{rgb}$. The temporal modeling module includes layers of Residual Attention Blocks constructed from Multi-head self-attention, Layer Norm, QuickGELU and MLP:
\begin{equation}
\mathcal{F}^{rgb-t} = \Phi_{TEMP}(\{\mathcal{F}^{rgb}(1), ..., \mathcal{F}^{rgb}(T)\})
\end{equation}
In terms of text embeddings $\mathcal{F}^{text} \in \mathbb{R}^{C \times d}$, a set of action categories are prepended with prompt template "a video of action" and then passed through the CLIP language encoder $\Phi_{CLIP-t}(.)$. For ZS video action recognition task, a mean pooling is commonly followed by $\mathcal{F}^{rgb-t}$ to get the aggregated video embedding, which is then used to find the highest matching score with $\mathcal{F}^{text}$. This cannot be naively applied in TAD because there are multiple events in individual video input. Therefore, we generate temporal semantic representation $\mathcal{S} \in \mathbb{R}^{T \times C}$ made up of visual and text features:
\begin{equation}
\mathcal{S} = \mathcal{F}^{rgb-t} \cdot (\mathcal{F}^{text})^{T}
\end{equation}
To recognize the category of an action proposal, we find the corresponding region on semantic embedddings and deduce the activity label from them. The embedding selection and aggregation on semantic representation are supported by dual-localization module. We analyze it in the next section.

\subsection{Dual Localization} \label{ssec:num3}
The backbone for dual-localization framework is motivated by Deformable DETR \cite{zhu2020deformable}, an encoder-decoder object detection model based on the Transformer. While the predecessor DETR \cite{carion2020end} takes long time to converge and its attention modules fails to generate high-resolution image feature maps, Deformable DETR requires remarkably less time to train and more importantly, achieves better performance at detecting small objects. This characteristic brings advantage in coping with the vagueness in action boundaries, temporal redundancy in long videos and short action detection. Initially, RGB frames are sampled into length $\mathcal{T}$ and pretrained 3D video encoder (e.g., I3D \cite{carreira2017quo} or TSP \cite{alwassel2021tsp}) extracts the video snippet features $\mathcal{F}^{3D} \in \mathbb{R}^{T \times l}$. A 1D convolutional is followed by to match video feature dimension with CLIP feature dimension. The encoder with $\mathcal{L}_E$ Transformer layers models the relations among video features via deformable attention modules and returns feature sequence $\mathcal{F}^{enc} \in \mathbb{R}^{T \times d}$ carrying temporal context. The decoding network consists of $\mathcal{L}_D$ deformable cross attention layers and receives the encoded feature sequence $\mathcal{F}^{enc}$ (served as key) and $\mathcal{N}_{q}$ learnable embedding queries $q \in \mathbb{R}^{d}$. For each query $q$, the decoder outputs an embedding $\mathcal{F}^{dec} \in \mathbb{R}^{d}$, which are later utilized by three prediction heads. 

\textbf{Boundary regression head} predicts the normalized action middle point and duration of an activity. The two variables are computed by applying a feed-forward network (FFN) then a sigmoid function into the output embedding:
\begin{equation}
\hat{Y}_b = \{ m,d \}= \operatorname{sigmoid}(FFN(\mathcal{F}^{dec}))
\end{equation}

\textbf{Actionness prediction head:} Conventionally, image object detection model uses classification score to rank the duplicate queries and pick the best bounding boxes. Since 3D encoded snippet feature does not comprehend strong discriminative power and adjacent frames of an action event share high similarity with the main ones, classification score is not a reliable ranking indicator for localization quality. Hence, we adopt the actionness score from TadTR \cite{liu2022end} to support proposal selection. In detail, the ROIAlign \cite{he2017mask} along with predicted boundaries $B$ is applied on the encoder output features $\mathcal{F}^{enc}$ to extract the small feature map within the action interval:
\begin{equation}
f^{RoI} =  RoI(\mathcal{F}^{enc},B)
\end{equation}
Subsequently, the aligned feature $f^{RoI}$ is fed to an FFN with sigmoid activation to regress the actionness score. Actionness score is effective because it guides the model to be more sensitive to the local features \cite{liu2022end}.

\textbf{Segment generator head} is the continuation of section \ref{ssec:num2}. To classify an event, we should attentively accumulate relevant semantic embeddings based on the action boundary. Using the start and end timestamps straight from the boundary regression head as binary segmentation is not optimal. For instance, the action "high jump" and "long jump" both comprise of striding and jumping. Equal attentions to these sub-actions could lead to classification uncertainty. Based on this observation and motivated by the mask formulation of MaskFormer \cite{cheng2021per}, we propose the proposal segment generator module. The semantic mask $\mathcal{M} \in \mathbb{R}^{T}$ is created by multiplying the video snippet feature $\mathcal{F}^{3D}$ with the prediction embedding $\mathcal{F}^{dec}$ via dot product:
\begin{equation}
\mathcal{M} =  \operatorname{sigmoid}(\mathcal{F}^{3D} \cdot \mathcal{F}^{dec})
\end{equation}
The semantic mask is kept flexible ($\mathcal{M} \in [0,1]$) instead of applying thresholding to binarize them. The region mask helps localize the events on the semantic embeddings and put more concentration on semantic units that provide decent classification clues. The final class of a action proposal is determined by follow:
\begin{equation}
\hat{Y}_{c} = \arg\max \frac{1}{T} \sum_{t=1}^T (\mathcal{M}_{(t)} \odot S_{(t)}) 
\end{equation}

\begin{figure}[t]
\centering
\includegraphics[scale=0.7]{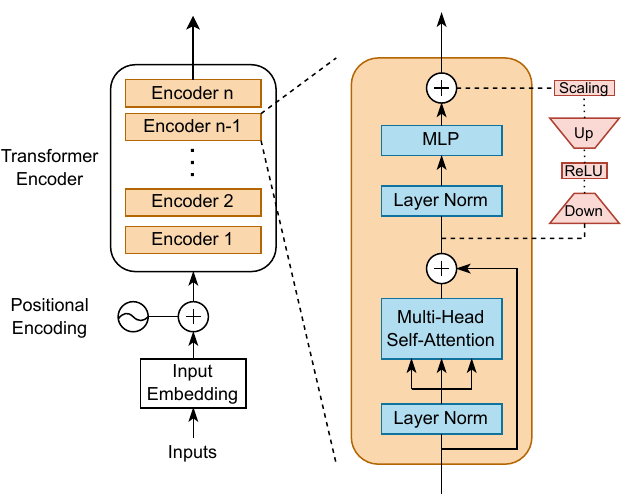}
\caption{Adapter-based finetuning mechanism in our ZEETAD.}
\label{fig:adaptformer}
\end{figure}

\subsection{Vision-Language Model Finetuning}
Fully finetuning on large-scale datasets has been well-known to instill the knowledge to downstream tasks. Considering the massive scale of the Transformer model, full finetuning takes tremendous amount of time and computation. Furthermore, full finetuning tends to overfit the downstream task and loses the generalization acquired in the large-scale pretraining stage, which is extremely harmful to ZS tasks. To address the above challenges, existing works \cite{zhou2022learning,jia2022visual,khattak2023maple} focused on parameter-efficient finetuning technique that adjusts minimum amount of learnable parameters and keep the entire backbone frozen. In this paper, we leverage AdaptFormer \cite{chen2022adaptformer} to implement our Adapter-based finetune the pretrained CLIP text encoder. As displayed in Fig.\ref{fig:adaptformer}, our Adapter-based finetune replaces the original MLP sub-layers in Transformer blocks with AdaptMLP, which is a bottleneck module parallel to the original one. During finetuning, only the added parameters are optimized and the entire encoder is frozen. The additional modules only occupied for 1.46\% of model parameters.

\subsection{Training and Inference}
\textbf{Training:} Following \cite{zhu2020deformable}, we include boundary refinement on each decoding layer and set dropout rate as zero within the transformer. Bipartite matching is used to find the lowest matching cost among targets to their corresponding predictions (one-to-one mapping) and the loss is computed on the matching pairs. While the matching cost regards the classification probabilities and the resemblance between target and output bounding boxes, the total training loss is defined as follow:
\begin{equation}
\mathcal{L} = \lambda_1 .\mathcal{L}_{cls} + \lambda_2 .\mathcal{L}_{bbox} + \lambda_3 .\mathcal{L}_{actionness}
\end{equation}
To maximize the alignment between the frame sequence embeddings and corresponding text embeddings, cross-entropy loss with temperature parameter $\tau$ is used:

\begin{equation}
\mathcal{L}_{cls} = -\sum_i \log\frac{\exp(f_i \cdot t_j/\tau)}{\sum_j \exp(f_i \cdot t_j/\tau)}
\end{equation}

$\mathcal{L}_{bbox}$ regresses the midpoint and the duration of the action proposal via L1 loss and the generalized IoU loss:

\begin{equation}
\mathcal{L}_{bbox} = \alpha_{1}.||b_{gt}-b_{i}||_1 + \alpha_{2}.gIoU(b_{gt},b_{i})
\end{equation}

Actionness score is supervised by the offset of the IoU between the predicted bounding box and its closest groundtruth segment. The higher overlap rate is equivalent to higher actionness score $a$, and vice versa. $\mathcal{L}_{actionness}$ is a L1 loss that minimizes the ranking loss of all action proposals:

\begin{equation}
\mathcal{L}_{actionness} = \sum || a_{i} - IoU(b_{gt},b_{i}) ||_1 
\end{equation}

\textbf{Inference}. For each video input, we have $\mathcal{N}_{q}$ proposal boundaries, labels, classification scores and actionness scores. The proposal labels are obtained by applying $argmax$ on the classification head's logits. The final proposal confidence score is obtained by multiplying classification scores and actionness scores. We gather the predictions in top-k order according to the confidence score. At last, Soft-NMS \cite{bodla2017soft} is applied to remove duplicate and low-quality predictions.

\begin{table*}[h]
\centering
\caption{Performance comparison between our ZEETAD with SOTA ZS TAD methods on ActivityNet-1.3 and THUMOS14. mAPs at different IOU thresholds of 0.3, 0.4, 0.5, 0.6, 0.7 and averaging (AVG) are reported.}
\begin{tabular}{c| c| c c c c c c| c c c c c}
\toprule
\multirow{2}{*}{\textbf{Train-Test split}} & \multirow{2}{*}{\textbf{Method}} & \multicolumn{6}{c|}{\textbf{THUMOS14}} & \multicolumn{4}{c}{\textbf{ActivityNet-1.3}}\\ \cline{3-12}
           &     & 0.3  & 0.4  & 0.5  & 0.6  & 0.7 & \textbf{AVG} & 0.5  & 0.75  & 0.95  & \textbf{AVG}\\
\midrule
 \multirow{5}{*}{75\%-25\%} & B-II \cite{nag2022zero} & 28.5 & 20.3 & 17.1 & 10.5 & 6.9 & 16.6 & 32.6 & 18.5 & 5.8   & 19.6 \\
           & B-I \cite{nag2022zero} & 33.0 & 25.5 & 18.3 & 11.6 & 5.7 & 18.8 & 35.6 & 20.4 & 2.1 & 20.2 \\
           & Eff-Prompt \cite{ju2022prompting} & 39.7 & 31.6 &23.0 & 14.9 & 7.5 & 23.3 & 37.6 & 22.9 & 3.8 & 23.1 \\
           & STALE \cite{nag2022zero} & 40.5 & 32.3 & 23.5 & 15.3 & 7.6 & 23.8 & 38.2 & 25.2 & \textbf{6.0} & 24.9 \\
           & ZEETAD & \textbf{61.4} & \textbf{53.9} & \textbf{44.7} & \textbf{34.5} & \textbf{20.5} & \textbf{43.2} & \textbf{51.0} & \textbf{33.4} & 5.9 & \textbf{32.5} \\
\midrule
\multirow{5}{*}{50\%-50\%} & B-II \cite{nag2022zero} & 21.0 & 16.4 & 11.2 & 6.3 & 3.2 & 11.6 & 25.3 & 13.0 & 3.7 & 12.9 \\
           & B-I \cite{nag2022zero} & 27.2 & 21.3 & 15.3 & 9.7 & 4.8 & 15.7 & 28.0 & 16.4 & 1.2 & 16.0 \\
           & Eff-Prompt \cite{ju2022prompting} & 37.2 & 29.6 & 21.6 & 14.0 & 7.2 & 21.9 & 32.0 & 19.3 & 2.9 & 19.6 \\
           & STALE \cite{nag2022zero} & 38.3 & 30.7 & 21.2 & 13.8 & 7.0 & 22.2 & 32.1 & 20.7 & \textbf{5.9} & 20.5 \\
           & ZEETAD & \textbf{45.2} & \textbf{38.8} & \textbf{30.8} & \textbf{22.5} & \textbf{13.7} & \textbf{30.2} & \textbf{39.2} & \textbf{25.7} & 3.1 & \textbf{24.9} \\

\bottomrule
\end{tabular}

\label{tab:comparison}
\end{table*}

\section{Experimental Results}

\subsection{Dataset and Metrics}
We conduct experiments on THUMOS14 \cite{idrees2017thumos} and ActivityNet-1.3 \cite{caba2015activitynet} datasets. THUMOS14 collects videos from 20 sports action classes, containing 200 and 213 videos for training and testing, respectively. ActivityNet-1.3 contains 200 classes of daily activities with the total of 19994 videos. We follow previous literature \cite{ju2022prompting} to adopt two validation splits (75\% seen training classes - 25\% novel testing classes and 50\% seen training classes - 50\% novel testing classes) for zero-shot scenarios. To ensure the legitimate generalization, the final results are averaged on 10 random splits. We compare to only existing methods that comply with this evaluation scheme. Following existing TAD and ZS TAD methods, the mean average precision (mAP) at different IOU thresholds is reported as main evaluation metrics. 

\subsection{Implementation Details}
Similar to other TAD methods, available 3D video features are used as the input of the localization module. The selection of video features is centered around the best performance on action boundary localization. On THUMOS14, we utilize the two stream I3D \cite{carreira2017quo} video encoder while on ActivityNet-1.3,TSP features \cite{alwassel2021tsp} is adopted. In terms of ActivityNet-1.3, we resize the video features to standard length $\mathcal{T}$ of 100 by linear interpolation. For THUMOS14, we follow previous papers \cite{lin2021learning} to divide the long video features into windows of length 128 with overlap rate of 0.75 for both training and testing. Regarding the ZS classification module, the model version of CLIP model is ViT-B/16. The input frames are picked from the middle frame of a snippet and thus, the length of the CLIP video features are equal to length $\mathcal{T}$ of the 3D video encoded features. We set $\mathcal{L}_E$ = 4, $\mathcal{L}_D$ =4, and $\mathcal{N}_q$ = 10 on ActivityNet and $\mathcal{N}_q$ = 40 on THUMOS. All parameters of the network including add-in adapters are learnable except the CLIP vision encoder and text encoder. Adam optimizer with learning rate of $10^{-4}$ and batch size of 16 are set as the model hyperparameters. ZEETAD is trained for 30 epochs on both datasets. The threshold for Soft-NMS is 0.3. 

\begin{table*}[h]
\centering
\caption{The effectiveness of different finetuning methods on two data split settings on THUMOS14 dataset. mAPs at different IOU thresholds of 0.3, 0.4, 0.5, 0.6, 0.7 and averaging (AVG) are reported.}
\begin{tabular}{l| c c c c c c| c c c c c c}
\hline
 & \multicolumn{6}{c|}{\textbf{75\%-25\%}} & \multicolumn{6}{c}{\textbf{50\%-50\%}}  \\
\cline{2-13}
 & 0.3 & 0.4 & 0.5 & 0.6 & 0.7 & \textbf{AVG} & 0.3 & 0.4 & 0.5 & 0.6 & 0.7 & \textbf{AVG} \\
 \midrule
w/o Finetuning & 52.3 & 46.6 & 38.9 & 29.2 & 19.1 & 37.2 & 39.2 & 34.2 & 28.1 & 20.6 & 13.2 & 27.1  \\ 
Text Prompt Tuning & 53.5 & 47.0 & 37.5 & 27.0 & 16.9 & 36.4 & 37.1 & 31.5 & 25.12 & 18.4 & 11.6 & 24.7 \\
Adapter & \textbf{61.4} & \textbf{53.9} & \textbf{44.7} & \textbf{34.5} & \textbf{20.5} & \textbf{43.2} & \textbf{45.2} & \textbf{38.8} & \textbf{30.8} & \textbf{22.5} & \textbf{13.7} & \textbf{30.2} \\
\bottomrule
\end{tabular}

\label{tab:finetuning}
\end{table*}

\subsection{Main Result}
Table \ref{tab:comparison} displays results of ZEETAD and other existing methods on open-set scenarios of TAD. We also include two baseline experiments (B-I and B-II) from STALE \cite{nag2022zero}. Overall, we achieve the state-of-the-art results at nearly all IoU thresholds on two datasets as well as two ZS settings. Specifically, on THUMOS14, in terms of average mAP, we surpass the second-best method, STALE, by 19.4\% on 75\%-25\% data split and by 8.0\% on 50\%-50\% data split. The large gain margins are also perceived on ActivityNet-1.3. ZEETAD achieves average mAPs of 32.5\% and 24.9\% on the mentioned dataset. The reported performance validates the effectiveness of our approach towards the problem of temporal action detection generalization.

\subsection{Ablation Study}
We implement experiments on THUMOS14 dataset and evaluate the options of constituent components and model structures in this section.

\textbf{Vision-Language model finetuning:} Different methods of CLIP model finetuning are investigated in Table \ref{tab:finetuning}. The average mAPs for not applying finetuning techniques are 37.2\% and 27.1\% on 75\%-25\% and 50\%-50\%, respectively. We implement Text Prompt Tuning based on CoOp \cite{zhou2022learning} and use the appended learnable context length of 16. We observe the performance drop in both dataset splits compared to the one not using TPT. The performance gap between TPT and baseline in 50\% setting is bigger than the 75\% setting, which could be an overfitting problem. According to \cite{zhou2022conditional}, the learned context is prone to overfitting on the seen classes and cannot be generalizable to unseen classes. By integrating adapter \cite{chen2022adaptformer}, we improve the average mAP by 6\% and 3.1\% on two data settings.

\begin{table}[t]
\centering
\caption{The effect of encoded feature utilization for localization and classification modules on 75\%-25\% setting on THUMOS14 dataset. mAPs at different IOU thresholds of 0.3, 0.5, 0.7 and averaging (AVG) are reported.}
\begin{tabular}{c| c| c c c c}
\toprule
Localizer & Classifier & 0.3 & 0.5 & 0.7 & \textbf{AVG} \\
\midrule
CLIP & CLIP & 47.3 & 29.7 & 11.5 & 29.7 \\
I3D & I3D & 43.4 & 33.0 & 16.9 & 31.5\\
I3D & CLIP & \textbf{61.4} & \textbf{44.7} & \textbf{20.5} & \textbf{43.2} \\ 
\bottomrule
\end{tabular}
\vspace{-4mm}
\label{tab:inputfeature}
\end{table}

\textbf{Encoded feature utilization:} To demonstrate the input problem of STALE, our network takes in uni-visual feature encoder backbone for both localizer and classifier. Experimented on 75\%-25\% setting, Table 3 describes the placement of two types of visual features (CLIP and I3D) for localization and classification modules. We can easily observe in Table \ref{tab:inputfeature} that using encoded features with no regard to their dedicated purpose results in big drop in performance. CLIP visual encoder have no capability to convey the temporal context and using online temporal learning model is trivial compared to 3D pretrained video encoder like I3D. This could be derived from the fact that using I3D on both branches yields better results.

\textbf{Component effectiveness:} We assess the importance of supporting components proposed in our method in Table \ref{tab:component}. Although the model is still able to elaborate the ZS TAD task if these components are omitted, their improvement cannot be denied. Action proposal confidence score is fused from classification score and actionness score. Without actionness prediction, the average mAP is reduced by 0.9\%. Temporal modeling module helps inject temporal context into CLIP encoded frame features, creating a temporal-coherent semantic representation. Directly generating semantic embeddings from frame embeddings decreases the performance from 43.2\% to 41.7\%. Segment generator is in charge of providing the soft masks of semantic embeddings corresponding to action proposals. Should it be excluded, we need to segment the semantic representation via action boundaries, which is analogous to hard (binary) mask prediction. The average mAP is declined to 33.4\% if segment generator is not employed, making it the key component in our network.

\begin{table}[!b]
\centering
\vspace{-4mm}
\caption{Ablation study of proposed components on 75\%-25\% setting on THUMOS14 dataset. mAPs at different IOU thresholds of 0.3, 0.5, 0.7 and averaging (AVG) are reported. Actionness Prediction, Temporal Modeling and Segment Generator are denoted as AP, TM and SG, respectively.}
\resizebox{\linewidth}{!}{%
\begin{tabular}{l|c c c c}
\toprule
 & 0.3 & 0.5 & 0.7 & \textbf{AVG} \\
\midrule
ZEETAD & 61.4 & 44.7 & 20.5 & 43.2 \\ 
- AP & 62.6 (+1.2) & 43.9 (-0.8) & 19.7 (-0.8) & 42.3 (\textbf{-0.9})\\
- TM & 59.6 (-1.8) & 43.4 (-1.3)& 20.3 (-0.2) & 41.7 (\textbf{-1.5})\\
- SG & 47.6 (-13.8) & 34.5 (-10.2) & 17.0 (-3.5) & 33.4 (\textbf{-9.8})\\
\bottomrule
\end{tabular}
}

\label{tab:component}
\end{table}

\begin{table}[h]
\centering
\caption{Analysis of model structure on 75\%-25\% setting on THUMOS14 dataset. mAPs at different IOU thresholds of 0.3, 0.4, 0.5, 0.6, 0.7 and averaging (AVG) are reported.}
\begin{tabular}{l|c c c c c c }
\toprule
 & 0.3 & 0.4 & 0.5 & 0.6 & 0.7 & \textbf{AVG} \\
\midrule
Two-stage  & 40.3 & 36.1 & 28.9 & 21.4 & 14.2 &  28.2\\
One-stage & \textbf{61.4} & \textbf{53.9} & \textbf{44.7} & \textbf{34.5} & \textbf{20.5} & \textbf{43.2} \\ 
\bottomrule
\end{tabular}
\label{tab:modelorganization}
\vspace{-4mm}
\end{table}

\textbf{Model structure:} To ascertain the advantage of one-stage ZS TAD over two-stage counterpart, ZEETAD is compared to its variant in Table \ref{tab:modelorganization}. To convert ZEETAD to two-stage framework, the localization branch is trained separately and the predicted action proposals are given to gather the semantic embeddings of the classification branch. This setup resembles Efficient-Prompt \cite{ju2022prompting}. We obtain the final average result of 28.2\%, which is much lower than the result of one-stage setup. Compared to the result of \cite{ju2022prompting} on the same setting, we still achieve better performance. This could be attributed to the stronger class-agnostic proposal detection backbone (Deformable DETR \cite{zhu2020deformable}) and also the more efficient deep prompt tuning technique.

\section{Conclusion}

In this work, we propose the Transformer-based end-to-end framework for Zero-shot Temporal Action Detection model, titled ZEETAD. The model is designed as a one-stage TAD method that unite the localization and classification tasks. Large-scale Vision-Language pretrained model plays an important role to empower the zero-shot classification capability. ZEETAD revises the CLIP zero-shot mechanism in image recognition and transforms it into video action proposal classification. Concurrently, the localization branch enhances the action recognition rate by segmenting the appropriate semantic embeddings with strong discriminative features. In addition, an efficient finetuning method known as adapter is integrated to CLIP text encoder. Adapter helps augment the text embeddings so that they could increase the matching score with their corresponding video embeddings. The experimental results on THUMOS14 and ActivityNet-1.3 verified the effectiveness of our approach and ZEETAD significantly outperforms existing methods.

\section*{Acknowledgment}
This material is based upon work supported by the National Science Foundation (NSF) under Award No OIA-1946391 RII Track-1, NSF 1920920 RII Track 2 FEC, NSF 2223793 EFRI BRAID, NSF 2119691 AI SUSTEIN, NSF 2236302.

{\small
\bibliographystyle{ieee_fullname}
\bibliography{egbib}
}

\end{document}